\documentclass[letterpaper, 10 pt, conference]{ieeeconf}  % Comment this line out if you need a4paper

\IEEEoverridecommandlockouts                              % This command is only needed if 
\title{\LARGE \bf
Dy3DGS-SLAM: Monocular 3D Gaussian Splatting SLAM for Dynamic Environments
}

\author{ Mingrui Li$^{1}$\textsuperscript{†},  Yiming Zhou$^{2,3,5}$\textsuperscript{†}, Hongxing Zhou$^{4}$, Xinggang Hu$^{1}$, Florian Roemer$^{3}$, \\ Hongyu Wang$^{1*}$, Ahmad Osman$^{2,3}$ % <-this % stops a space
\thanks{\textsuperscript{†} Both authors contributed equally to the paper.
        }%
\thanks{$^{*}$ Corresponding author.
        }%
\thanks{$^{1}$ Mingrui Li, Xinggang Hu and Hongyu Wang are with the School of Information and Communication Engineering, Dalian University of Technology, Dalian, 116024, China
        {\tt\small 2905450254@mail.dlut.edu.cn}}%
\thanks{$^{2}$ Yiming Zhou and Ahmad Osman are with the School of Engineering Sciences, Saarland University of Applied Sciences, Saarbruecken, 66117, Germany
        {\tt\small yiming.zhou@htwsaar.de, ahmad.osman@htwsaar.de}}
\thanks{$^{3}$ Florian Roemer, Yiming Zhou and Ahmad Osman are with Fraunhofer Institute for Nondestructive Testing, Saarbruecken, 66123, Germany
     {\tt\small Florian.Roemer@izfp.fraunhofer.de}}
\thanks{$^{4}$ Hongxing Zhou is with the College of Information Science and Technology, Beijing University of Chemical Technology, Beijing, 100029, China
        {\tt\small hongxingzhou@mail.buct.edu.cn}}
\thanks{$^{5}$ Yiming Zhou is with the Faculty of Science and Engineering, Laval University, Quebec, GEV 0A6, Cananda
        {\tt\small yiming.zhou.1@laval.ca}}
}

\usepackage{cite}
\usepackage{graphicx}
\usepackage{verbatim}
\usepackage{amsmath}  % For advanced math typesetting
\usepackage{amssymb}  % For additional math symbols
\usepackage{bm}       % For bold math symbols
\usepackage[capitalise]{cleveref}

\usepackage{color,soul}
\definecolor{light-gray}{gray}{0.8}

\usepackage{amsmath,amssymb,amsfonts}
\usepackage{algorithmic}
\usepackage{graphicx}
\usepackage{textcomp}
\usepackage{caption}
\usepackage{subcaption}
\usepackage{colortbl}
\usepackage{multirow}
\usepackage{array}
\usepackage{booktabs}
\usepackage[table]{xcolor} % Required for custom colors
\usepackage{float}
\usepackage{arydshln}

\begin{document}

\maketitle
\thispagestyle{empty}
\pagestyle{empty}

%%%%%%%%%%%%%%%%%%%%%%%%%%%%%%%%%%%%%%%%%%%%%%%%%%%%%%%%%%%%%%%%%%%%%%%%%%%%%%%%
% Update
% 光流提取部分,根据后端高斯体rendering，对前端进行修饰，结果陈同学有。NeRF采样指导高斯体生成，用高斯体去跟踪，（最主要的问题 前后端分离太严重），分割动态物体不准，比如光线动态
% 得用dytanvo 跑他的数据集，2页实验
% 7天后写完方法，14天后完成结果写完

% 后端提取光流，每20 frames，图像扔到前端进行mapping，特点是3d本身有一部分抗动态，novel view可消除一部分动态
% 主要创新点: 不依赖点云和内存小，去除动态伪影干扰，后端提取光流可以减少瞬态干扰; 多视图合成实现图像增强（变亮） ;yolo有预定义;直接法效果不好

% Dataset: TUM? + ?

%%%%%%%%%%%%%%%%%%%%%%%%%%%%%%%%%%%%%%%%%%%%%%%%%%%%%%%%%%%%%%%%%%%%%%%%%%%%%%%%
\begin{abstract}
% 只说gs，目前方法在后端提取光流
%The current Simultaneous Localization and Mapping (SLAM) methods based on NeRF or 3D Gaussian Splatting have shown impressive results in reconstructing ideal static 3D scenes. However, they perform poorly in tracking and reconstruction when facing more challenging dynamic environments, such as real-world scenes involving dynamic elements. Although some NeRF-based SLAM methods have attempted to address these dynamic challenges, they rely on RGB-D inputs, and there is a lack of methods that work with pure RGB input.
%To address these challenges, we introduce Dy3DGS-SLAM, the first 3D Gaussian Splatting SLAM method for dynamic scenes using monocular RGB input. For tracking, our method first acquires dynamic object masks through an optical flow estimation system, then combines them with a monocular depth estimation system to obtain merged masks and recover scale. This allows us to remove dynamic objects from non-predefined scenes, enabling dense frame-to-frame mapping. For rendering, we prune the Gaussians generated by pixels with dynamic masks, while applying a scale regularizer to avoid Gaussian artifacts. We impose additional photometric, geometric, and uncertainty losses on the proxy depth to improve rendering accuracy. Experimental results show that our method achieves state-of-the-art (SOTA) tracking and rendering results in dynamic environments, while also being competitive with or outperforming RGB-D methods.

Current Simultaneous Localization and Mapping (SLAM) methods based on Neural Radiance Fields (NeRF) or 3D Gaussian Splatting excel in reconstructing static 3D scenes but struggle with tracking and reconstruction in dynamic environments, such as real-world scenes with moving elements. Existing NeRF-based SLAM approaches addressing dynamic challenges typically rely on RGB-D inputs, with few methods accommodating pure RGB input.
To overcome these limitations, we propose Dy3DGS-SLAM, the first 3D Gaussian Splatting (3DGS) SLAM method for dynamic scenes using monocular RGB input. To address dynamic interference, we fuse optical flow masks and depth masks through a probabilistic model to obtain a fused dynamic mask. With only a single network iteration, this can constrain tracking scales and refine rendered geometry. Based on the fused dynamic mask, we designed a novel motion loss to constrain the pose estimation network for tracking. In mapping, we use the rendering loss of dynamic pixels, color, and depth to eliminate transient interference and occlusion caused by dynamic objects. Experimental results demonstrate that Dy3DGS-SLAM achieves state-of-the-art tracking and rendering in dynamic environments, outperforming or matching existing RGB-D methods.

\end{abstract}

\section{INTRODUCTION}

Recently, dense SLAM systems expressed through NeRF \cite{mildenhall2021nerf} or 3DGS \cite{kerbl20233d} have attracted significant attention. These systems have achieved photo-realistic rendering results in static scenes and are gradually expanding towards large-scale or challenging scenarios. However, a practical issue for SLAM systems is evident: the real world contains a large number of dynamic objects, and current NeRF or 3DGS-based SLAM systems \cite{sucar2021imap, zhu2022nice, sandstrom2023point, tosi2024nerfs, zhou2024evaluating} perform poorly in addressing this challenge. Another issue that has gained attention is how to achieve better results without relying on RGB-D sensors and using only monocular RGB input, which is considered a more accessible sensor and a solution with greater potential.

Although some NeRF-based methods have attempted to address dynamic objects, such as DN-SLAM\cite{ruan2023dn}, DDN-SLAM\cite{li2024ddn}, NID-SLAM\cite{xu2024nid}, and RoDyn-SLAM\cite{jiang2024rodyn}, they often rely on predefined dynamic priors or heavily depend on depth priors to determine dynamic object masks, making them unsuitable for environments with only monocular RGB input. Furthermore, due to the limitations of NeRF \cite{mildenhall2021nerf} representation, there are constraints on rendering accuracy, often resulting in severe rendering artifacts. 3DGS-based SLAM systems, such as SplaTAM \cite{keetha2024splatam}, Photo-SLAM \cite{huang2024photo}, and MonoGS \cite{matsuki2024gaussian}, perform well in static environments, but they tend to encounter tracking failures and mapping errors in dynamic scenes.

Therefore, we propose Dy3DGS-SLAM, the first RGB-only 3DGS-SLAM system designed for dynamic environments. 
%\textcolor{red} 
{We utilize optical flow to obtain dynamic masks without relying on predefined moving objects, though these masks can be noisy in regions with uniform textures or fast motion. To address this, we incorporate monocular depth estimation, providing complementary spatial cues, especially for occlusions and depth discontinuities. We then propose a depth-regularized mask fusion strategy that combines the strengths of both modalities, mitigating individual limitations and producing more precise, robust dynamic masks.}

% 在跟踪方面，我们将估计的深度和融合掩模引入到运动损失中，在姿态估计网络中有效地恢复了尺度和位姿，获得了更准确的跟踪结果。
For tracking, we incorporate the estimated depth and fused mask into the motion loss, effectively recovering scale and pose in the pose estimation network, resulting in more accurate tracking outcomes.
% For rendering, 为了解决瞬态干扰和遮挡，我们基于动态像素的颜色和深度对动态高斯进行惩罚，相比于基线方法，我们的渲染伪影显著减小，几何精度显著提升。
In terms of rendering, to address transient interference and occlusion, we penalize dynamic Gaussians based on the color and depth of dynamic pixels. Compared to baseline methods, our approach significantly reduces rendering artifacts and greatly improves geometric accuracy.
% For rendering, we perform pruning of the Gaussian volumes corresponding to the pixels of the fusion mask and impose geometric, photometric, and uncertainty losses on the estimated depth. 
% To ensure that the Gaussian volumes at the edges of the dynamic mask do not produce elongated artifacts, we apply regularization loss to the edge depths. 
In summary, our method has the following contributions:

\begin{itemize}
    \item We propose Dy3DGS-SLAM, the first RGB-only 3DGS-SLAM system for dynamic environments, capable of robust tracking and high-fidelity reconstruction in dynamic environments.
    \item %\textcolor{red} 
    {We propose a mask fusion method that accurately covers dynamic objects by combining motion cues from optical flow with geometric consistency from depth estimation.} Based on the fused mask, we introduce novel motion and rendering losses to effectively mitigate dynamic object interference in tracking and rendering.

    % \item \textbf{Tracking}: pose estimation based on fusion mask and estimated depth. \textbf{Rendering}: pruning based on the fusion mask and rendering loss on the estimated depth.
    % 我们提出了融合掩码方法，无需多次网络更新，即可准确覆盖动态物体；基于融合掩码，我们提出了新颖的运动损失和渲染损失，可以解决动态物体对跟踪和渲染的干扰。
    \item Our results on three real-world datasets demonstrate that our method achieves better tracking and rendering performance compared to baseline methods.
\end{itemize}

%################################################################################
\section{Related Work}

Vision-based SLAM systems are essential technologies for addressing mapping challenges in robotics and scene reconstruction in VR/AR applications \cite{zhu2023demonstration, song2023going, song2024looking}.
In real-world scenarios, dynamic objects pose a significant challenge to visual SLAM systems. There has been extensive exploration in the traditional visual SLAM field to tackle the interference caused by dynamic objects. Recently, deep learning-based methods have gained attention, mainly falling into two categories. 
One type focuses on semantic prior-based segmentation, represented by systems such as DS-SLAM\cite{yu2018ds}, OVD-SLAM\cite{he2023ovd}, and SG-SLAM\cite{cheng2022sg}. These methods utilize deep learning frameworks to recognize semantic information and then use epipolar geometry or depth constraints to identify dynamic points and remove them. The second type relies on optical flow estimation methods, such as FlowFusion\cite{zhang2020flowfusion}, DeflowSLAM \cite{ye2022deflowslam}, and DytanVO\cite{shen2023dytanvo}, using deep learning-based optical flow estimation frameworks to estimate camera poses. However, these methods often lack dense and stable reconstruction and fail to accurately recover depth or obtain precise dynamic object masks under monocular conditions.

With the advent of NeRF and 3DGS showing high-fidelity reconstruction and fast rendering capabilities in 3D reconstruction, there has been a growing interest in RGB-D SLAM systems. However, these systems often perform poorly in real-world dynamic environments. Some NeRF-based RGB-D SLAM systems have explored this problem. For instance, DN-SLAM\cite{ruan2023dn} uses optical flow estimation to remove dynamic points and employs the Instant NGP-based \cite{muller2022instant} rendering framework for view synthesis, although it struggles with artifacts. NID-SLAM \cite{xu2024nid} leverages an optical flow estimation system to obtain dynamic masks and completes background reconstruction, but its tracking accuracy is limited, and the rendering process lacks sufficient constraints. Rodyn-SLAM \cite{jiang2024rodyn} uses a sliding window optical flow estimation method to acquire motion masks and proposes specific rendering losses, but it heavily depends on prior depth information provided by the sensors. Our approach utilizes the advantages of 3DGS for representation while addressing the reliance on depth sensors through a depth estimation system. By using optical flow estimation, we generate more accurate motion masks, enabling the reconstruction of static scenes effectively.

\section{Method}
% 1. 后到前
% 2. 初始化像dytan VO，我们知道不准，但是误差很小 并且后端可以进行矫正，对每frame 高斯体进行rendering和光流，用nerf降低瞬态干扰初始化，所以不是点云

%按照ngmslam改，
%参考rodynslam写，肯定要写剪枝

% new：
% 图加optical flow
% 主打单目，然后mapping加个depth estimation，来指导前端重新跟踪
% depthanything 估计写后端

\begin{figure*}
  \centering
      \includegraphics[width=0.9\linewidth, height=7cm]{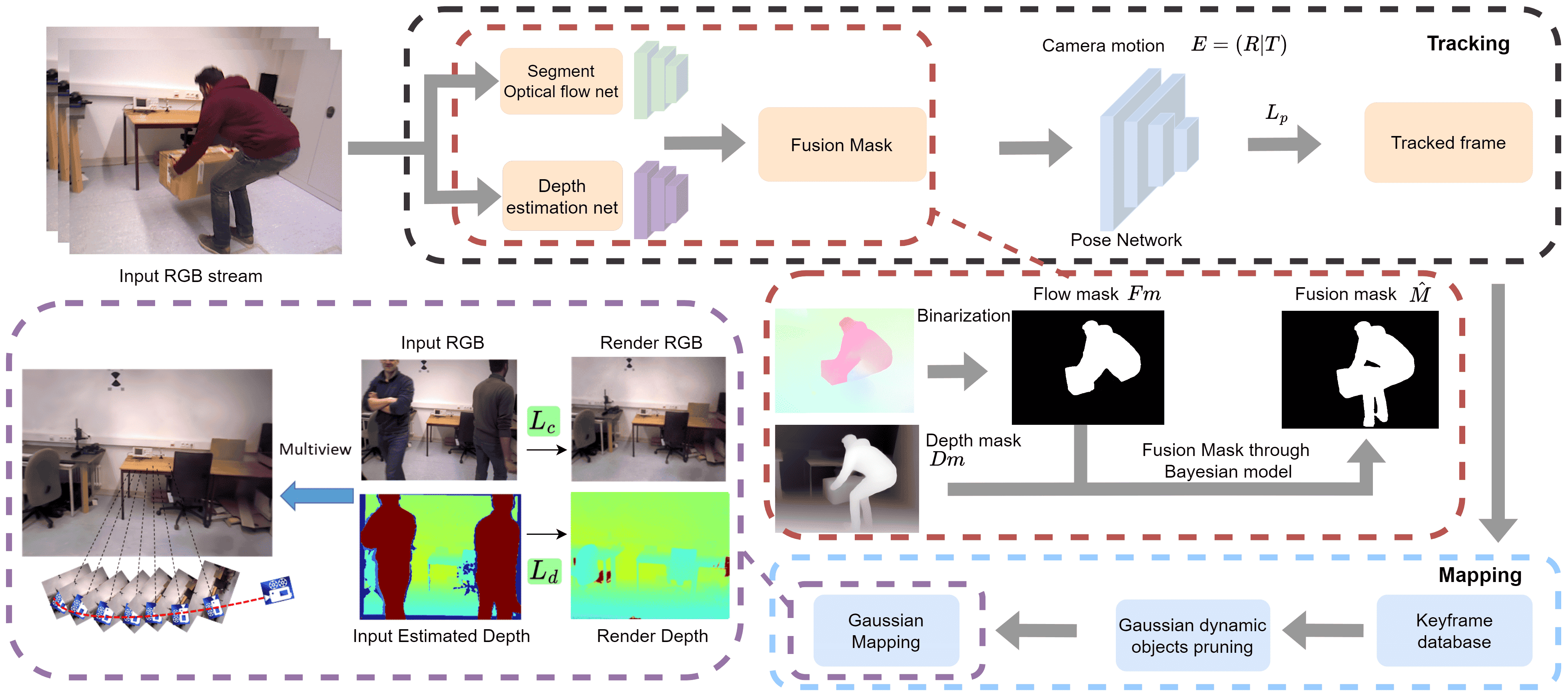}
    \caption{\textbf{Pipeline of Our Network:} Our system workflow consists of two main threads: tracking and mapping. In the tracking thread, we use a segmentation optical flow network and a depth estimation network to generate the estimated motion optical flow mask and depth map mask. By applying a conditional probability approach, we create a fused mask \(\hat{M}\). This fused mask is subsequently input into the pose estimation network to determine the estimated pose. 
In the mapping thread, we utilize the fused mask \(\hat{M}\) along with keyframes to construct the map. We impose color and depth penalties on the Gaussians corresponding to the moving pixels identified by the fused mask, which ultimately results in a multi-view rendering outcome.
}
    \label{fig:pipeline}
% \vspace{-5mm}
\end{figure*}

Our system pipeline is shown in \cref{fig:pipeline}. 
In \cref{sec:dynamic_mask_fusion}, we address the problem of fusing the dynamic mask obtained from optical flow with the depth map estimated from monocular input, resulting in an accurate dynamic fusion mask. 
In \cref{sec:tracking_monocular_dynamic}, we propose the motion estimation network and introduce a motion loss incorporating depth estimation, enabling the network to iteratively refine accurate camera poses. 
In \cref{sec:gaussian_rendering_monocular}, we penalize the Gaussians corresponding to pixels labeled as dynamic and apply an additional rendering loss based on monocular depth to optimize the scene details. Finally, we synthesize a static scene using multi-view consistency.

\subsection{Dynamic Mask Fusion}
\label{sec:dynamic_mask_fusion}
Our tracking network includes an optical flow estimation module, a depth estimation module, a mask fusion method, and a pose estimation module. Assume that we have two consecutive undistorted images, $I_t$ and $I_{t+1}$, as input, and the output is the relative camera motion $E = (R|T)$, where $T \in \mathbb{R}^3$ represents the 3D translation and $R \in \text{SO}(3)$ represents the 3D rotation.

To detect optical flow anomalies caused by dynamic objects, we employ a lightweight U-Net \cite{ronneberger2015u} motion segmentation network. It takes the original input frames as input and binarizes the result, setting all optical flow within the mask area to zero, and finally obtains the corresponding optical flow values $F$ and the optical flow mask $F_m$. However, this estimation is relatively coarse and static areas are easily mis-segmented, affecting the accuracy of camera pose estimation. Therefore, we introduce additional depth supervision using the DepthanythingV2 \cite{yang2024depth} estimation network to provide estimated depth, obtaining the corresponding depth mask $D_{m}$.

% Although we have obtained both the dynamic mask from optical flow estimation and the dynamic mask from depth, these two individual masks are incomplete. Therefore, 
% 尽管我们获取了光流估计网络生成的动态掩码，但是掩码存在错误估计的情况，因此我们需要利用深度掩码对光流掩码进行校正。
Although we obtained dynamic masks generated by the optical flow estimation network, there are instances of incorrect mask estimation. Therefore, we need to use the depth mask to correct the optical flow mask.
We combine depth and optical flow information using conditional probability to obtain a more accurate fusion mask \(\hat{M} \) for determining dynamic regions.

First, we aggregate optical flow pixels to separate multiple potential moving objects. We use the K-means clustering algorithm to segment the motion pixels. Let the set of pixels in the motion region be \( P_{\text{dynamic}} = \{ p \mid M(p) = 1 \} \). We divide these motion pixels into \( k \) clusters, with each cluster representing an independent dynamic object. The goal of clustering is to minimize the sum of squared pixel distances within each cluster. The clustering objective function is defined as follows:
\begin{equation}
\min_{\mu_1, \mu_2, \dots, \mu_k} \sum_{i=1}^{k} \sum_{p \in C_i} \| p - \mu_i \|^2,
\end{equation}
where \( C_i \) represents the set of pixels assigned to the \( i \)-th cluster, and \( \mu_i \) is the center of the \( i \)-th cluster, defined as:
\begin{equation}
\mu_i = \frac{1}{|C_i|} \sum_{p \in C_i} p.
\end{equation}

Through the clustering result, we can separate moving objects and ensure the correctness of the subsequent fusion mask process. Finally, we obtain the set of moving objects $N = \{ N_1, N_2, \dots, N_k \}$, where $N_i$ represents the pixel set of the $i$-th moving object.
We perform a separate depth probability search for each moving object to achieve mask fusion. Since the depth map and optical flow mask are independent, we propose a Bayesian model for estimating the probability:
%$P(D_{map}, F_d | M(p) = 1)$ represents the joint likelihood of the depth map and optical flow mask when the pixel belongs to a moving object. $P(D_{map}, F_d)$ is the normalization term in all cases.
\begin{equation}
P(D_{m}, F_m \mid M(p)) = P(D_{m} \mid M(p)) \cdot P(F_m \mid M(p)).
\end{equation}

We binarize the mask \( M(p) \) by setting it to 1, where \( P(M(p) = 1 \mid D_{m}, F_m) \) represents the posterior probability that a pixel belongs to a moving object. 
\( P(D_{m} \mid M(p) = 1) \) is the likelihood of observing the depth map \( D_{m} \) given that the pixel is part of a moving object, and \( P(F_m \mid M(p) = 1) \) is the likelihood of observing the optical flow mask \( F_m \) under the same assumption.
\( P(M(p) = 1) \) denotes the prior probability that a pixel is part of a moving object.

Finally, by combining all known information and calculating the posterior probability for each pixel, a new motion mask for each pixel is obtained:
\begin{equation}
\hat{M}_i = 
\begin{cases} 
1 & \text{if } P(M(p) = 1 | D_{m}, F_m) > T \\
0 & \text{otherwise}
\end{cases}, %\tag{4}
\end{equation}
where \( T \) is the pixel probability threshold, set to 0.95 in our method.

The final fused mask for all moving objects is given by \( \hat{M} = M(N_1) \cup M(N_2) \cup \cdots \cup M(N_i) \).
Our method does not require additional network iterations and is more generalized, allowing it to handle multiple moving objects simultaneously without the need for scene-specific parameter adjustments.

\subsection{Tracking for Monocular Dynamic Scenes}
\label{sec:tracking_monocular_dynamic}

Unlike other visual odometry systems \cite{wang2021tartanvo}, our systems can provide accurate pixel depth constraints due to the use of estimated depth information. This is significant in addressing pose estimation errors caused by depth ambiguity. We obtain the static depth mask $M_{ds} = \{ p \in \hat{M} \, | \, M(p) = 0 \}$ through the fusion of dynamic masks. The static depth mask is applied to the optical flow map, and the corresponding scale factor $S_n$ is fused to obtain the formula:
\begin{equation}
\tilde{F} = F \cdot M_{ds} \cdot S_n.
\end{equation}

In this formula, $F$ is the optical flow map, $M_{ds}$
filters the residual mask of untrustworthy dynamic regions, and $S_n$ provides accurate scale information for the remaining static regions. Through this method, the network can apply more reliable depth information to the static areas, avoiding interference from dynamic regions while correcting scale errors in monocular estimation.

We fuse the acquired static mask $M_{ds}$ with the optical flow mask $\tilde{F}$ and input them into the network to update the pose.
To achieve a more accurate iterative pose estimation process, we introduce a camera motion loss $\mathcal{L_M}$, adjusting the estimated pose distance from the ground truth. The loss function with the introduced scale constraint is expressed as:
\begin{equation}
\mathcal{L_M} = \frac{\hat{T}}{\max(|\hat{T} \cdot S_n| , \varepsilon)} - \frac{T}{\max(|T \cdot S_n| , \varepsilon)} + (\hat{R} - R) \cdot M_{ds},
\end{equation}
where $S_n$ is used to adjust the scale of the translation vector to align it with the true scale value. We perform pose updates within a pose estimation network based on ResNet50 \cite{he2016deep}, following the training design in TartanVO\cite{wang2021tartanvo}. The network jointly optimizes optical flow loss $\mathcal{L_O}$, motion segmentation loss $\mathcal{L_U}$
 , and camera motion loss $\mathcal{L_M}$, which incorporates depth masking and scale constraints. DytanVO\cite{shen2023dytanvo} improves camera pose estimation and dynamic mask segmentation through three iterations, but the final results still have limitations. In contrast, our method requires only a single iteration without incurring additional computational costs. The comprehensive tracking loss function is formulated as:
\begin{equation}
\mathcal{L_P} = \lambda_1 \mathcal{L_O} + \lambda_2 \mathcal{L_U} + \mathcal{L_M},
\end{equation}
where $\lambda_1$ and $\lambda_2$ are weights that control the different loss terms, ensuring that the network can balance the tasks of optical flow, motion segmentation, and pose estimation during training.

For more accurate pose estimation, we generate a keyframe every 10 frames and create a keyframe group consisting of at least 4 keyframes, applying local Bundle Adjustment optimization to correct accumulated errors.

\subsection{Gaussian Rendering for Monocular Dynamic Scenes}
\label{sec:gaussian_rendering_monocular}

In 3DGS \cite{kerbl20233d}, an explicit point-based scene representation is optimized. Each 3D Gaussian is parameterized by a set of 3D attributes, including position, opacity, scale, and rotation. The Gaussian ellipsoid is characterized by a full 3D covariance matrix \(\boldsymbol{\Sigma}\), which is defined (normalized) in world space. The Gaussian function is defined as:
\begin{equation}
g(\mathbf{x}) = o \exp\left(-\frac{1}{2} \mathbf{x}^T \boldsymbol{\Sigma}^{-1} \mathbf{x}\right),
\boldsymbol{\Sigma} = \boldsymbol{R} \boldsymbol{S} \boldsymbol{S}^T \boldsymbol{R}^T,
\end{equation}
where \(\boldsymbol{\Sigma}\) is the covariance matrix, \(o \in [0, 1]\) represents the opacity value, \(S\) is the scale matrix, and \(R\) is the rotation matrix.

We use 3D Gaussian ellipsoids to render 2D images through splatting techniques, as described in \cite{kopanas2021point, yifan2019differentiable}. In the camera coordinate system, the covariance matrix \(\boldsymbol{\Sigma}'\) is formulated as:
\begin{equation}
\boldsymbol{\Sigma}' = \boldsymbol{J} \boldsymbol{W} \boldsymbol{\Sigma} \boldsymbol{W}^T \boldsymbol{J}^T,
\end{equation}
where \(\boldsymbol{W}\) represents the viewing direction, and \(\boldsymbol{J}\) is the Jacobian matrix of the affine approximation of the projection transformation. For each pixel, the color and opacity of all Gaussian ellipsoids are computed and blended using the following formula:
\begin{equation}
C = \sum_{i \in N} \boldsymbol{c}_i g_i \prod_{j=1}^{i-1} \left(1 - g_j\right),
\end{equation}
where \(\boldsymbol{c}_i\) represents the color of the \(i\)-th Gaussian ellipsoid. Additionally, we propose a similar formula for depth rendering:
\begin{equation}
D = \sum_{i=1}^n d_i g_i \prod_{j=1}^{i-1} \left(1 - g_j\right),
\end{equation}
where \(d_i\) is the z-axis depth of the center of the \(i\)-th 3D Gaussian.

Since our method is based on keyframe multi-view rendering, each Gaussian \(g_i\) is associated with a keyframe that anchors it to the map \(G\). For the Gaussians produced by pixels marked as dynamic, we set their depth to infinity to perform pruning. However, this may cause artifacts that are hard to remove, so we apply photometric loss \(L_{c}\) and depth loss \(L_{d}\) to the pixel masking process to eliminate the artifacts' impact.

The new photometric loss is:
\begin{equation}
L_{c} = \lambda_d \cdot \frac{N_d}{N_{pi}} \left| C_k - C_k^{gt} \right| + \lambda_s \cdot \frac{N_{pi} - N_d}{N_{pi}} \left| C_k - C_k^{gt} \right|,
\end{equation}
where \(\lambda_d\) is the penalty factor for dynamic pixel masks, \(\lambda_s\) is the penalty factor for static pixel masks, \(N_{pi}\) is the number of pixels in each keyframe, and \(N_d\) represents the number of pixels corresponding to the dynamic mask.

The new depth loss is:
\begin{equation}
L_{d} = \lambda_t \cdot \frac{D_d}{D_{pi}} \left| D_k - D_k^{e} \right| + \lambda_m \cdot \frac{D_{pi} - D_d}{D_{pi}} \left| D_k - D_k^{e} \right|,
\end{equation}
where \(\lambda_t\) is the penalty factor for dynamic depth masks, \(\lambda_m\) is the penalty factor for static depth masks, \(D_{pi}\) is the number of pixels corresponding to the estimated depth in each keyframe, \(D_d\) represents the depth corresponding to the dynamic mask, and \(D_k^{e}\) represents the depth generated by monocular estimation.

The final rendering loss function \(L_G\) is:
\begin{equation}
L_G = L_{c} + \lambda \cdot L_{d},
\end{equation}
% where \(Kframe\) contains the keyframe set in the local window, and \(\lambda\) is a hyperparameter set to 1.
where \(\lambda\) is a hyperparameter set to 1.

\section{Experimental Results}
%coslam 补充是nidslam结果，
% rodynslam结果是eslam补充
% Bonn+TUM
% refusion+dynaSLAM+rodynslam, must be discussed in related work!
\subsection{Experimental Details and Metrics}
{\bf{Datasets and Implementation details.}}
We evaluated our method on three public datasets from the real world: the TUM RGB-D dataset \cite{sturm12iros}, AirDOS-Shibuya dataset \cite{qiu2022airdos} and the BONN RGB-D dynamic dataset \cite{palazzolo2019iros}, all of which capture real indoor environments.
We conducted our SLAM experiments on a desktop equipped with a single RTX 3090 Ti GPU. We present results from our multiprocess implementation designed for real-time applications. Consistent with the 3DGS framework, time-critical rasterisation and gradient computation are implemented using CUDA.

%We run our SLAM on a desktop with a single RTX 3090Ti GPU. We present results from our multiprocess implementation aimed at real-time applications. As with 3DGS, time-critical rasterisation and gradient computation are implemented using CUDA. 
% We  set \( \lambda_{color} = 0.9\) for all RGB-D experiments and \( \lambda_{ssim} = 0.5\).

{\bf{Metrics and Baseline Methods.}}
To evaluate camera tracking accuracy, we report the Root Mean Square Error (RMSE) of the Absolute Trajectory Error (ATE) for keyframes. For runtime performance and network iteration speed, we measure frames per second (FPS) and milliseconds (ms), respectively. GPU usage is assessed in megabytes (MB).
We compare our Dy3DGS-SLAM method against traditional dynamic SLAM approaches, such as ORB-SLAM3 \cite{campos2021orb}, Droid-SLAM \cite{teed2021droid},
DynaSLAM \cite{jiang2024rodyn}, DytanVO\cite{shen2023dytanvo} and ReFusion \cite{palazzolo2019iros}, as well as state-of-the-art NeRF-based methods utilizing RGB-D sensors, including NICE-SLAM \cite{zhu2022nice}, ESLAM \cite{johari2023eslam}, Co-SLAM \cite{wang2023co}, and NID-SLAM \cite{xu2024nid}. Furthermore, we consider SplaTAM \cite{keetha2024splatam}, which is based on 3DGS.

\begin{figure*}[h]
	\centering
	\includegraphics[scale=0.52]{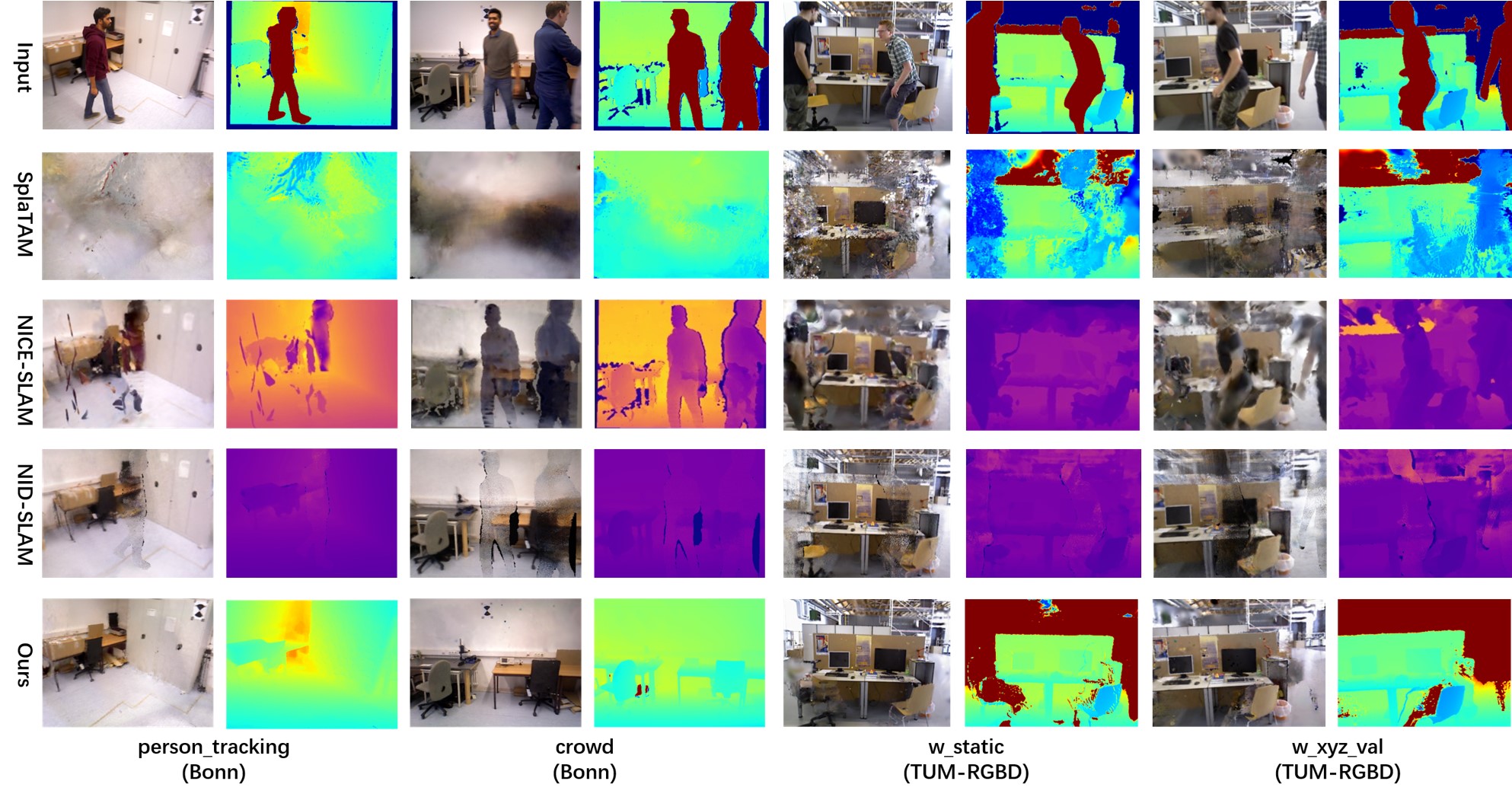}
	\caption{Visual comparison of the reconstructed meshes on the BONN and TUM RGB-D datasets. Our results are more complete and accurate without the dynamic object floaters.}
	\label{fig:visual_result}
\end{figure*}

\definecolor{LightGreen}{rgb}{0.85, 1, 0.85}
\definecolor{MediumGreen}{rgb}{0.6, 1, 0.6}
\definecolor{DarkGreen}{rgb}{0.2, 0.8, 0.2}
\definecolor{LightYellow}{rgb}{1, 1, 0.85}
\definecolor{MediumYellow}{rgb}{1, 1, 0.6}
\definecolor{DarkYellow}{rgb}{1, 1, 0.3}
\definecolor{first}{RGB}{192, 226, 202}
\definecolor{second}{RGB}{227, 237, 186}
\definecolor{third}{RGB}{255, 247, 188}
%%%%%%%%%%%%%%%%%%%%%%%% Tracking table Bonn %%%%%%%%%%%%%%%%%%%%%%%% 
\begin{table*}[htbp!]
	\centering
	\caption{Tracking performance on \textit{BONN-RGB-D} (ATE RMSE ↓ [cm]). The best results are highlighted as \colorbox{first}{\bf{first}}, \colorbox{second}{\bf{second}} and \colorbox{third}{\bf{third}}.}
	\setlength{\tabcolsep}{8pt}
	\begin{tabular}{lcccccccccccc}
		\toprule
		\textbf{Methods}  & ballon & ballon2 & ps\_track & ps\_track2 & mv\_box1 & mv\_box2 & \textbf{Avg.} \\
		\rowcolor{gray!20} % Light grey background for the entire row
		\textit{Traditional SLAM methods} &  &  &  &  &  &  &  &  \\
		ORB-SLAM3 \cite{campos2021orb}  & 5.8  & 17.7 & 70.7  & 77.9 & \colorbox{first}{\bf{3.1}} & \colorbox{first}{\bf{3.7}} & 29.8 \\
		Droid-SLAM \cite{teed2021droid}  & 5.4 & 4.6 & 21.3 & 46.0 & 8.9 & 5.9 & 15.4 \\
		DynaSLAM \cite{bescos2018dynaslam}  & \colorbox{first}{\bf{3.0}} & \colorbox{third}{\bf{3.0}}  & \colorbox{third}{\bf{6.1}} & \colorbox{second}{\bf{7.8}} & \colorbox{second}{\bf{4.9}} & \colorbox{third}{\bf{3.9}} & \colorbox{second}{\bf{4.8}} \\
		ReFusion \cite{palazzolo2019iros}  & 17.5 & 25.4 & 28.9 & 46.3 & 30.2 & 17.9  & 27.7 \\
            DytanVO\cite{shen2023dytanvo}  & 6.3 & 3.1 & \colorbox{first}{\bf{3.5}} & \colorbox{third}{\bf{9.0}} & 5.5 & 6.1  & \colorbox{third}{\bf{5.6}} \\
		\midrule
		
		\rowcolor{gray!20} % Light grey background for the entire row
		\textit{NeRF or 3DGS based SLAM methods} &  &  &  &  &  &  &  &  \\
		NICE-SLAM \cite{zhu2022nice}  & 80.3 & 66.8 & 54.9 & 45.3 & 21.2 & 31.9 & 44.1 \\
		Co-SLAM \cite{wang2023co}  & 28.8 & 20.6 & 61.0 & 59.1 & 38.3 & 70.0 & 46.3 \\
		ESLAM \cite{johari2023eslam} & 22.6 & 36.2 & 48.0 & 51.4 & 12.4 & 17.7 & 31.4  \\
		NID-SLAM \cite{xu2024nid} &\colorbox{second}{\bf{3.7}}  &\colorbox{second}{\bf{2.8}}  &10.0  &14.7  &6.9  &4.2  &7.1   \\
		SplaTAM \cite{keetha2024splatam} &231.9  &126.4  &27.8  &53.1  &63.7  &73.6  &96.1   \\
		Dy3DGS-SLAM (Ours)  &\colorbox{third}{\bf{4.5}}  &\colorbox{first}{\bf{1.9}}  &\colorbox{second}{\bf{5.6}}  &\colorbox{first}{\bf{6.4}}  & \colorbox{third}{\bf{5.2}} &\colorbox{second}{\bf{3.8}}  &\colorbox{first}{\bf{4.5}}  \\ 
		\bottomrule
	\end{tabular}
	\label{tab:bonn_rgbd_results}
\end{table*} %2.92.6

\subsection{Evaluation on TUM and Bonn RGB-D}
{\bf{Tracking.}}
As shown in \cref{tab:tum_rgbd_results}, we present results for three highly dynamic sequences, one mildly dynamic sequence, and two static sequences from the TUM dataset \cite{sturm12iros}. Thanks to our proposed dynamic mask fusion method, our system demonstrates advanced tracking performance compared to RGB-D-based methods and is even competitive with traditional SLAM methods.
Furthermore, we evaluated the tracking performance on the more complex and challenging BONN dataset \cite{palazzolo2019iros}, as illustrated in \cref{tab:bonn_rgbd_results}. Even in these more complicated and large-scale scenarios, our method achieved superior performance. 
Our method outperforms all other approaches, with NID-SLAM\cite{xu2024nid} being the only one achieving results close to ours. Additionally, our method demonstrates superior performance compared to traditional methods. This highlights our dynamic mask fusion can effectively remove the dynamic objects and enhance the tracking process.

{\bf{Mapping.}}
To comprehensively evaluate the performance of our proposed system in dynamic scenes, we analyze the results from a qualitative perspective. We compare the rendered images with ground truth poses obtained from the generated Gaussian map, using the same viewpoint as other methods. Four challenging sequences were selected: $crowd$ and $person\_tracking$ from the BONN dataset, and $f3\_walk\_xyz\_val$ and $f3\_walk\_static$ from the TUM RGB-D dataset. As shown in \cref{fig:visual_result}, our method shows significant advantages in geometric and texture details, especially in reducing artifacts.
Notably, our approach is based on a monocular system and has been validated on two real-world datasets, demonstrating its capability to accurately record dynamic scenes with just a simple camera. This highlights the potential of our method to effectively track and reconstruct indoor environments, making it a valuable tool for applications where depth sensors may not be available.

%%%%%%%%%%%%%%%%%%%%%%%% Tracking table TUM %%%%%%%%%%%%%%%%%%%%%%%% 
\begin{table*}[htbp!]
	\centering
	\caption{Tracking performance on \textit{TUM-RGB-D} (ATE RMSE ↓ [cm]). ``-" denotes the absence of mention. “X” denotes the tracking failures. The best results are highlighted as \colorbox{first}{\bf{first}}, \colorbox{second}{\bf{second}} and \colorbox{third}{\bf{third}}.}
	\setlength{\tabcolsep}{8pt}
	\begin{tabular}{lcccccccccccc}
		\toprule
		\multirow{2}{*}{\textbf{Methods}}  & \multicolumn{4}{c}{\textbf{High Dynamic}} & \multicolumn{2}{c}{\textbf{Low Dynamic}} & \multirow{2}{*}{\textbf{Avg.}} \\
		\cmidrule(lr){2-5} \cmidrule(lr){6-7} 
		& f3/wk\_xyz & f3/wk\_hf & f3/wk\_st & f3/st\_hf & f3/st\_xyz & f1/st\_rpy & \\
		\midrule
		%\multicolumn{3}{l}{\textbf{Traditional SLAM methods}} \\
		\rowcolor{gray!20} % Light grey background for the entire row
		\textit{Traditional SLAM methods} &  &  &  &  &  &  &  &  \\
		ORB-SLAM3 \cite{campos2021orb}  & 28.1  & 30.5 &\colorbox{third}{\bf{2.0}}  & \colorbox{first}{\bf{2.6}} & 2.2 & \colorbox{first}{\bf{2.8}} & 11.5 \\
		DVO-SLAM \cite{kerl2013dense}  & 59.7 & 52.9 & 21.2 & 6.2 & \colorbox{third}{\bf{2.1}} & \colorbox{second}{\bf{3.0}} & 23.2 \\
		DynaSLAM \cite{bescos2018dynaslam}  & \colorbox{first}{\bf{1.7}} & \colorbox{first}{\bf{2.6}}  & \colorbox{first}{\bf{0.7}} & \colorbox{second}{\bf{2.8}} & \colorbox{first}{\bf{1.6}} &5.1  &\colorbox{first}{\bf{2.7}}\\
		ReFusion \cite{palazzolo2019iros}  & 9.9 & 10.4  &\colorbox{second}{\bf{1.7}} & 11.0 & - & -  & 8.3 \\
            DytanVO\cite{shen2023dytanvo}  & 8.7 & 9.8 & 9.5 & 14.7 & 12.4 & 12.3  & 11.2 \\
		\midrule
		
		\rowcolor{gray!20} % Light grey background for the entire row
		\textit{NeRF or 3DGS based SLAM methods} &  &  &  &  &  &  &  &  \\
		NICE-SLAM \cite{zhu2022nice}  & 113.8 & X & 137.3 & 93.0 & 43.9 & 65.6 & 90.6 \\
		Co-SLAM \cite{wang2023co}  & 51.8 & 105.1 & 49.5 & 4.7 & 9.3 & {8.9} & 36.3 \\
		ESLAM \cite{johari2023eslam} & 45.7 & 60.8 & 93.6 & 3.6 & 8.6 & 9.2 & 34.5  \\
		NID-SLAM \cite{xu2024nid} &\colorbox{third}{\bf{6.4}}  &\colorbox{third}{\bf{7.1}}  &6.2  &10.9  &7.5  &8.6  &\colorbox{third}{\bf{7.8}}   \\
		SplaTAM \cite{keetha2024splatam} &41.3  &72.6  &45.7  &75.9  & 32.8 &40.5  &53.2  \\
		Dy3DGS-SLAM (Ours)  &\colorbox{second}{\bf{5.8}}  & \colorbox{second}{\bf{7.0}} &6.5   &\colorbox{third}{\bf{3.4}}  &\colorbox{second}{\bf{2.0}} & \colorbox{third}{\bf{3.8}} & \colorbox{second}{\bf{4.7}}\\ 
		\bottomrule
	\end{tabular}
	\label{tab:tum_rgbd_results}
\end{table*}

%%%%%%%%%%%%%%%%%%%%%%%%%%%%%%%%% some results

\begin{figure}[htbp!]
  \centering
  \includegraphics[width=0.4\textwidth, height=6cm]{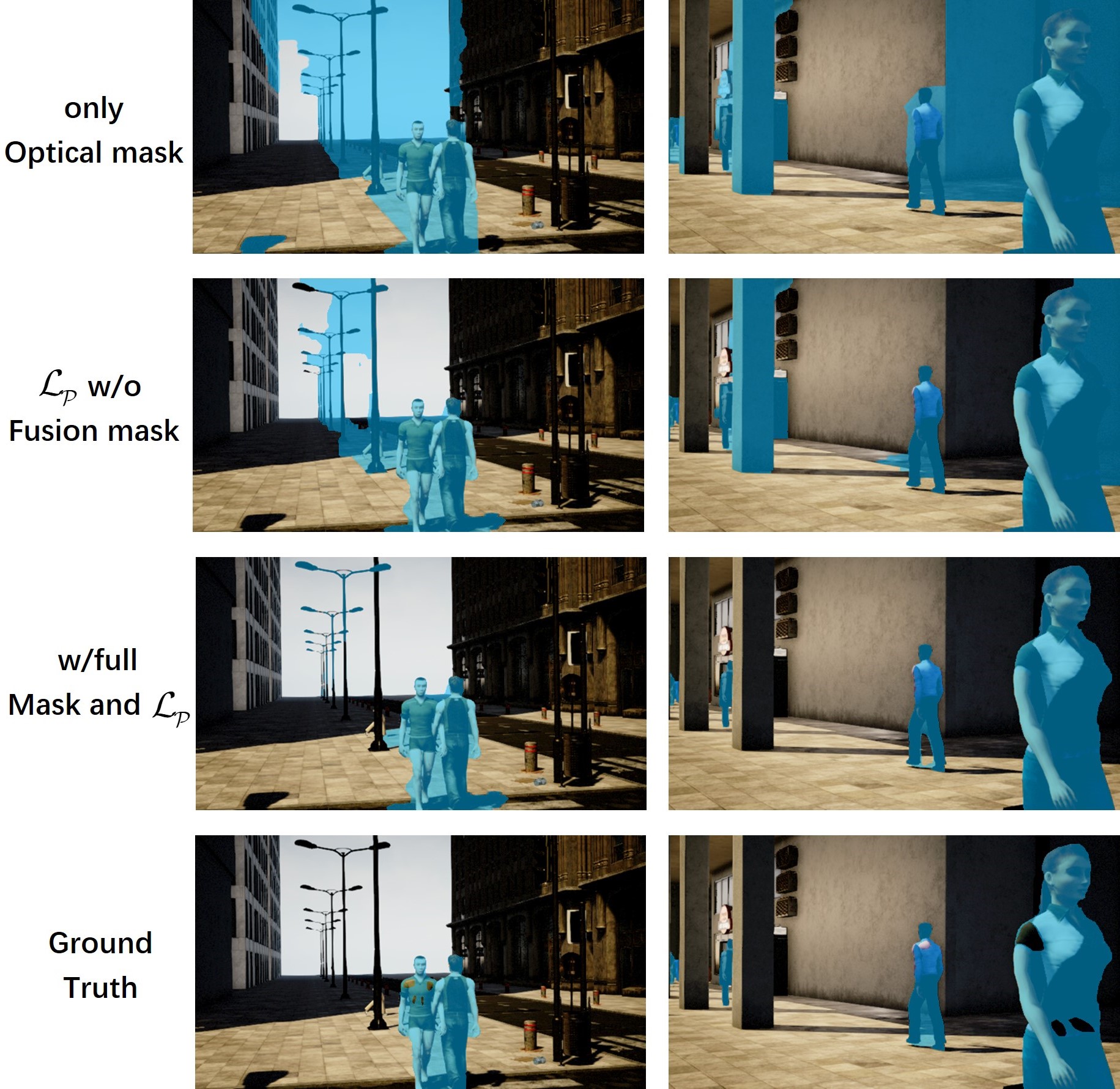} % adjust dimensions
  \caption{Evaluation of Tracking Network Loss Methods}
  \label{fig:example}
  \vspace{-15pt}
\end{figure}

%%%%%%%%%%%%%%%%%%%%%%%% Reconstructing table BONN %%%%%%%%%%%%%%%%%%%%%%%% 

\section{Ablation study}
\subsection{Fusion Mask Strategy Evaluation}
% 为了评估我们系统中提出的方法的有效性，我们对AirDOS Shibuya数据集中的五个场景进行了消融研究，所有结果都是五个实验的平均值，如\cref{tab:sablation_tracking-1}所示。我们计算平均绝对轨迹误差（ATE），以评估每种方法对整体系统性能的影响。结果表明，所有提出的方法都有助于改善相机跟踪。相比于只采用光流估计策略，我们的方法在ATE提升了60.52%，这表明，将光流掩模与深度掩模集成可以有效增强姿态估计。
% 而使用分而治之的姿态优化方法可以有效地提高相机跟踪的鲁棒性和准确性。
To evaluate the effectiveness of the proposed methods in our system, we conduct ablation studies on five scenes from the AirDOS-Shibuya dataset, with all results being the average of five experiments, as shown in \cref{tab:ablation_tracking-1}. We calculate the average Absolute Trajectory Error (ATE) to assess the impact of each method on the overall system performance. The results in \cref{tab:ablation_tracking} indicate that all the proposed methods contribute to improved camera tracking. Compared to using only the optical flow estimation strategy, our method improves ATE by 60.52\%, indicating that integrating the optical flow mask with the depth mask can effectively enhance pose estimation.
% This demonstrates that integrating the optical flow mask with the depth mask enhances pose estimation, while using a divide-and-conquer approach for pose optimization effectively boosts the robustness and accuracy of camera tracking.

\subsection{Evaluation of Mask Fusion Accuracy}
 We compared the mask results obtained with different loss functions $\mathcal{L_P}$ as shown in \cref{fig:example}. When using only the optical flow mask or only the motion loss similar to DytanVO, significant mask estimation errors appeared. Our strategy achieved the best results, closely approaching the ground truth.

%We conducted an ablation study of our tracking loss function on the AirDOS-Shibuya dataset. We compared the mask results obtained with different loss functions $\mathcal{L_P}$ as shown in \cref{fig:example}. When using only the optical flow mask or only the motion loss similar to DytanVO, significant mask estimation errors appeared. Our strategy achieved the best results, closely approaching the ground truth.

\subsection{Operating Speed and Network Performance Evaluation}
%We tested the system’s tracking, mapping, and network update speed on the AirDOS-Shibuya dataset. Compared to DytanVO and the state-of-the-art (SOTA) 3DGS-based SLAM system SplaTAM, our method achieves the best balance between runtime and performance. This is mainly because our system does not use a multi-iteration network update strategy.

We evaluated the system's tracking, mapping, and network update speed using the AirDOS-Shibuya dataset as shown in \cref{tab:ablation_tracking}. In comparison to DytanVO and the state-of-the-art 3DGS-based SLAM system SplaTAM, our approach demonstrates a superior balance between runtime efficiency and performance. This advantage primarily stems from the absence of a multi-iteration network update strategy in our system.

\begin{comment}
\begin{table}[h]
	\centering
	\caption{Title.}
	\begin{tabular}{l|c|c|c}
		\toprule
		& \textbf{Standing I} & \textbf{RoadCrossing III} & \textbf{RoadCrossing VII} \\ \midrule
		\textbf{1 iter}  & 0.0649             & 0.1666                   & 0.3157                   \\
		\textbf{2 iter}  & 0.0315             & 0.0974                   & 0.0658                   \\
		\textbf{3 iter}  & 0.0327             & 0.0608                   & 0.0660                   \\
		\textbf{Finetuned} & 0.0384             & 0.0631                   & 0.0531                   \\ 
		\bottomrule
	\end{tabular}
\end{table}
\end{comment}

\begin{comment}
\begin{table}[h!]
    \centering
    \caption{Ablation Study of the Proposed Method in Our Systems.}
    \begin{tabular}{lcccc}
        \toprule
        & \textbf{w/o Flow} & \textbf{w/o Depth} & \textbf{w/o Edge} & \textbf{RoDyn-SLAM} \\ 
        \midrule
        \textbf{ATE RMSE (m) $\downarrow$} & 0.3089 & 0.1793 & 0.2056 & \textbf{0.1354} \\ 
        \textbf{STD (m) $\downarrow$} & 0.1160 & 0.0739 & 0.0829 & \textbf{0.0543} \\ 
        \bottomrule
    \end{tabular}
\end{table}
\end{comment}

\begin{table}[h!]
    \centering
    \caption{Ablation study of the proposed method in our systems. The best result is highlighted as \colorbox{first}{\bf{first}}.}
    \begin{tabular}{cccc}
        \toprule
        Optical Flow & Depth & Optical Flow+Depth & ATE RMSE (cm) $\downarrow$ \\
		\midrule
        \textcolor{green}{\checkmark} & \textcolor{red}{\texttimes} & \textcolor{red}{\texttimes} & 7.6\\
        \textcolor{red}{\texttimes} & \textcolor{green}{\checkmark} & \textcolor{red}{\texttimes} & 94.8\\
        \textcolor{green}{\checkmark} & \textcolor{green}{\checkmark} & \textcolor{green}{\checkmark} & \colorbox{first}{\bf{3.0}} \\
        \bottomrule
    \end{tabular}
    \label{tab:ablation_tracking-1}
\vspace{-2mm}
\end{table}

\begin{table}[h!]
    \centering
    \caption{Operating speed and network performance. The best result is highlighted as \colorbox{first}{\bf{first}}.}
    \resizebox{\linewidth}{!}{ % 让表格自适应页面宽度

    \begin{tabular}{ccccc}
        \toprule
 Method   &       Tracking (FPS) $\uparrow$ & Mapping (ms) $\uparrow$ & Network update (ms) $\downarrow$ & GPU memory (MB) $\downarrow$ \\
		\midrule
  DytanVO   &   10.5 & \textcolor{red}{\texttimes} & 32.9  & \colorbox{first}{\bf{7.6}}\\ 
   SplaTAM   & 3.8 & 390.4 & \textcolor{red}{\texttimes} & 14.6\\
   Ours   &       \colorbox{first}{\bf{17.0}} & \colorbox{first}{\bf{430.5}} & \colorbox{first}{\bf{10.3}} & 12.8 \\
        \bottomrule
    \end{tabular}
    }
    \label{tab:ablation_tracking}
\end{table}

% CONCLUSION, 三段式: 1.论文总结； 2.缺点和不足； 3.未来工作 以及 社区贡献

\section{CONCLUSIONS}
% We propose Dy3DGS-SLAM, the first 3D Gaussian SLAM method designed for dynamic scenes using monocular RGB input. This method generates dynamic object masks through optical flow estimation, then combines these with monocular depth estimation to create fused masks and recover scale, enabling the removal of dynamic objects and dense frame-to-frame mapping. To enhance the pose accuracy of network estimation, we optimized the loss function based on the fused mask, reducing performance degradation from multiple iterations. Additionally, to improve rendering performance, we applied extra photometric and depth losses to eliminate transient interference artifacts and enhance geometric accuracy. Experimental results show that Dy3DGS-SLAM achieves state-of-the-art tracking and rendering performance in dynamic environments, and demonstrates competitive or superior performance compared to existing RGB-D methods.
We propose Dy3DGS-SLAM, the first 3DGS-based SLAM method designed for dynamic scenes using monocular RGB input. This method first generates dynamic object masks through optical flow estimation, combining these masks with monocular depth estimation to create a fused mask and recover scale, accurately capturing dynamic object masks. To further improve pose accuracy, we optimized the loss function based on the fused mask, reducing the computational cost associated with multiple iterations. Additionally, to enhance rendering performance, we applied additional photometric and depth losses to eliminate transient interference artifacts and improve geometric accuracy. Experimental results demonstrate that, compared to baseline methods, Dy3DGS-SLAM achieves state-of-the-art tracking and rendering performance in dynamic environments. In the future, we will focus on applying this approach to mobile devices with lower computational costs.
% 我们提出了 Dy3DGS-SLAM，这是第一个使用单目 RGB 输入设计的适用于动态场景的 3D 高斯 SLAM 方法。该方法首先通过光流估计生成动态物体掩码，将这些掩码与单目深度估计相结合，创建融合掩码并恢复比例，能够准确获取动态物体掩码。然后为了进一步提高位姿精度，我们基于融合掩码优化了损失函数，减少了多次迭代带来的计算成本。此外，为了改进渲染性能，我们应用了额外的光度损失和深度损失，以消除瞬态干扰伪影并提高几何精度。实验结果表明，与基线方法相比，Dy3DGS-SLAM 在动态环境中实现了最先进的跟踪和渲染性能。在未来，我们将在更低计算成本的移动端应用方面展开研究。


%%%%%%%%%%%%%%%%%%%%%%%%%%%%%%%%%%%%%%%%%%%%%%%%%%%%%%%%%%%%%%%%%%%%%%%%%%%%%%%%

%%%%%%%%%%%%%%%%%%%%%%%%%%%%%%%%%%%%%%%%%%%%%%%%%%%%%%%%%%%%%%%%%%%%%%%%%%%%%%%%

%%%%%%%%%%%%%%%%%%%%%%%%%%%%%%%%%%%%%%%%%%%%%%%%%%%%%%%%%%%%%%%%%%%%%%%%%%%%%%%%

\bibliographystyle{IEEEtran}
\bibliography{ref}

\begin{comment}

\end{comment}

\end{document}